\documentclass{article}

\usepackage{arxiv}
\usepackage[utf8]{inputenc} 
\usepackage[T1]{fontenc}    
\usepackage{hyperref}       
\usepackage{url}            
\usepackage{booktabs}       
\usepackage{amsfonts}       
\usepackage{nicefrac}       
\usepackage{microtype}      
\usepackage{lipsum}
\usepackage{graphicx}
\graphicspath{ {./images/} }
\usepackage{amsmath}
\usepackage{amssymb}

\usepackage{tabularx}
\usepackage{multirow}
\usepackage{rotating}

\title{Multi-Agent-Based Simulation of Archaeological Mobility in Uneven Landscapes}

\author{
 Chairi Kiourt \\
  Athena" Research Centre\\
  Xanthi, Greece \\
  \texttt{chairiq@athenarc.gr} \\
   \And
 Vassilis Evangelidis \\
  "Athena" Research Centre\\
   Xanthi, Greece \\
  \texttt{vevangelidis@athenarc.gr} \\
  \And
 Dimitris Grigoropoulos \\
  German Archaeological Institute Athens\\
  Athens, Greece \\
  \texttt{dimitris.grigoropoulos@dainst.de} \\
}

\begin{document}
\date{} 

\pagestyle{fancy}
\fancyhf{}
\renewcommand{\headrulewidth}{0pt}
\fancyfoot[C]{\thepage}

\maketitle
\begin{abstract}
Understanding mobility, movement, and interaction in archaeological landscapes is essential for interpreting past human behavior, transport strategies, and spatial organization, yet such processes are difficult to reconstruct from static archaeological evidence alone. This paper presents a multi-agent-based modeling framework for simulating archaeological mobility in uneven landscapes, integrating realistic terrain reconstruction, heterogeneous agent modeling, and adaptive navigation strategies. The proposed approach combines global path planning with local dynamic adaptation, through reinforcment learning, enabling agents to respond efficiently to dynamic obstacles and interactions without costly global replanning. Real-world digital elevation data are processed into high-fidelity three-dimensional environments, preserving slope and terrain constraints that directly influence agent movement. The framework explicitly models diverse agent types, including human groups and animal-based transport systems, each parameterized by empirically grounded mobility characteristics such as load, slope tolerance, and physical dimensions. Two archaeological-inspired use cases demonstrate the applicability of the approach: a terrain-aware pursuit and evasion scenario and a comparative transport analysis involving pack animals and wheeled carts. The results highlight the impact of terrain morphology, visibility, and agent heterogeneity on movement outcomes, while the proposed hybrid navigation strategy provides a computationally efficient and interpretable solution for large-scale, dynamic archaeological simulations.. 
\end{abstract}


\section{Introduction}

Understanding movement, mobility, and interaction in archaeological landscapes remains a fundamental yet challenging problem in cultural heritage research \cite{BeaudryParno2013, Verhagen2012Transit, BRANTINGHAM2025104895}. Human and animal movement patterns \cite{Verhagen2019, White-2021}, transport strategies, and spatial interactions have played a critical role in shaping settlement dynamics, economic exchange, conflict, and daily life in past societies, but they are often difficult to reconstruct from static archaeological evidence alone \cite{Lake2014,Conolly2006}. Traditional approaches based on GIS cost-surface analysis\cite{QUIROS2024105988} and least-cost paths provide valuable insights into potential routes and accessibility, yet they typically assume static environments, homogeneous actors, and optimal behavior, overlooking social heterogeneity, adaptive decision-making, and dynamic interactions. As a result, many mobility-related processes—such as evasive movement, group-dependent behavior, transport under load, or responses to terrain and visibility constraints—remain insufficiently explored within conventional analytical frameworks.

Agent-based modeling (ABM)\cite{5429318} has emerged as a promising paradigm for addressing these limitations by explicitly representing autonomous entities with heterogeneous capabilities, goals, and behavioral rules interacting within spatially explicit environments \cite{Kiourt2017,Bonabeau2002,Railsback2019}. Despite its growing adoption in domains such as urban studies \cite{GONZALEZMENDEZ2021105110} and crowd simulation\cite{Sidiropoulos2020}, ABM has been only sparsely applied in archaeology and cultural heritage, particularly for modeling movement and transport in uneven, large-scale landscapes \cite{Lake2014,Kiourt2017,Sidiropoulos2021}. Moreover, integrating realistic terrain derived from real-world elevation data introduces significant computational challenges\cite{PAVLIDIS200793}, especially when dynamic obstacles, agent interactions, and frequent path recalculations are required. Classical global path-planning techniques \cite{LuDa2025}, while effective for static routing, can become computationally expensive and unsuitable for real-time or large-scale simulations when repeatedly recomputed on high-resolution terrain meshes \cite{LaValle2006,PAVLIDIS200793,10.1145/3611314.3615901}. These challenges motivate the need for hybrid navigation approaches that balance global optimality, local adaptability, and computational efficiency. As a result, many mobility-related processes—such as evasive movement, group-dependent behavior, transport under load, or responses to terrain and visibility constraints—remain insufficiently explored within conventional analytical frameworks. However, few approaches combine realistic terrain, heterogeneous agents, and adaptive navigation within a unified, scalable simulation framework for archaeological analysis.

In this work, we propose a multi-agent-based modeling framework for simulating archaeological mobility in uneven landscapes that combines global path planning with local adaptive navigation for dynamic navigation optimization. Our main contributions are fourfold:(i) the integration of real-world Digital Elevation Models into high-fidelity three-dimensional simulation environments suitable for large-scale agent-based analysis, preserving metrically accurate terrain constraints;(ii) a hybrid navigation strategy that combines global A* path planning with local Q-learning-based adaptation, enabling agents to respond efficiently to dynamic obstacles and interactions without costly global replanning;(iii) the explicit modeling of heterogeneous agent types, including human groups and animal-based transport systems, parameterized using empirically grounded mobility constraints such as slope tolerance, load, bodily capability, and physical characteristics; and (iv) a robust archaeological interpretative framework that situates simulation outcomes within established archaeological debates on mobility, refuge, and transport, treating agent-based simulations as exploratory tools for evaluating plausible movement scenarios rather than deterministic reconstructions of past behavior.
The framework is demonstrated through two archaeological case studies: the Roman-period fort of Kimmeria (outside modern Xanthi), examined as a terrain-aware pursuit–evasion scenario, and the renown sanctuary of Kalapodi at Central Greece (near Atalanti), investigated through comparative terrestrial transport simulations. Together, these case studies will hopefully provide empirical grounding for evaluating how terrain morphology and agent heterogeneity influence mobility outcomes under varying environmental and technological conditions.

The remainder of this paper is organized as follows. Section 2 presents the background of the study, covering archaeological mobility and movement, agent-based modeling, navigation and path-planning strategies, and digital terrain representation. Section 3 describes the proposed methodology, including  heterogeneous agent modeling, and the hybrid navigation framework that combines global path planning with local adaptation. Section 4 introduces the archaeological use cases and simulation setups, while Section 5 reports and analyzes the simulation results. Finally, Section 6 concludes the paper and outlines directions for future research.

\section{Background}
This section reviews the theoretical and methodological foundations relevant to this study. It outlines key concepts in archaeological mobility and movement, agent-based modeling, navigation and path-planning strategies, and digital terrain representation, providing the background necessary to situate the proposed simulation framework within existing research.

\subsection{Mobility and Movement in Archaeological Landscapes}
Mobility and movement \cite{Llobera2000UnderstandingMovement} are fundamental to the understanding of archaeological landscapes because they reveal how people interacted with space, resources, and one another over time. Landscapes are not static backdrops to human activity \cite{Snead2009Landscapes}; rather, they are dynamic environments continuously shaped through acts of movement such as walking, herding, trading, migrating, worshipping, or fleeing. Through movement, people created connections between places, transformed natural terrain into meaningful space, and embedded social, economic, and symbolic practices within the landscape \cite{Ingold1993Temporality, Tilley1994Phenomenology}.
Movement operates across multiple scales. At a local level, everyday mobility linked settlements to agricultural fields, water sources, and sanctuaries, reflecting patterns of land use, labor organization, and social interaction. At broader regional and interregional scales, movement structured trade networks, pilgrimage routes, and military campaigns, shaping long-distance connectivity and cultural exchange. While archaeological remains such as roads, paths, waystations, and artifact distributions (specifically pottery) provide valuable evidence for these processes, they often capture only fragmentary and indirect traces of mobility. As a result, movement must be reconstructed through interpretation and analytical modeling rather than observed directly.
Over the past decades, methodological advances such as Geographic Information Systems (GIS), network analysis \cite{Mills2017SNA,Brughmans2010Connecting}, and computational modeling have significantly enhanced our ability to study mobility \cite{Llobera2012LifePixel,Herzog2013LeastCost}. These approaches allow archaeologists to reconstruct potential routes, evaluate accessibility, and assess the costs and constraints of movement imposed by terrain and infrastructure \cite{Verhagen2019ModellingPathways}. At the same time, there is growing recognition that mobility is not purely functional or deterministic. Human movement is shaped by perception, bodily experience, social relationships, cultural norms, and decision-making under uncertainty. Understanding mobility therefore requires approaches that integrate spatial data with human behavior, enabling archaeologists to move beyond static representations and toward dynamic interpretations of how landscapes were experienced, negotiated, and transformed through movement.

\subsection{Agent-Based Modeling of Entities}

Agent-based modeling (ABM) has been widely adopted as a computational paradigm for studying complex systems composed of interacting, autonomous entities. By representing individuals as agents with distinct attributes and behavioral rules, ABMs enable the exploration of emergent phenomena arising from local interactions within spatially explicit environments \cite{Bonabeau2002,Railsback2019}. This modeling approach is particularly suitable for domains where heterogeneity, adaptation, and non-linear interactions play a central role.

Within archaeology and cultural heritage research, the use of agent-based modeling remains relatively limited, with only a small number of studies addressing mobility, transport, and interaction dynamics that are difficult to infer from static evidence alone \cite{Lake2014}. Nevertheless, multi-agent virtual environments allow the explicit representation of heterogeneous capabilities, intentions, and environmental constraints, supporting exploratory analyses of plausible behavioral scenarios rather than deterministic reconstructions \cite{Kiourt2017}. Recent studies have further demonstrated the applicability of ABM to movement-centric problems, including crowd simulation for crisis management and heritage-related scenarios \cite{Sidiropoulos2020}, as well as the integration of adaptive decision-making through reinforcement learning techniques \cite{Sidiropoulos2021}. These works highlight the value of combining agent heterogeneity, environment-aware navigation, and adaptive behavior to model realistic dynamics in complex cultural landscapes. Building on this body of research, the present work employs agent-based modeling to represent diverse human and transport entities operating under terrain, visibility, and interaction constraints.

\subsection{Global Path Planning and Local Navigation Adaptation}

Navigation in spatial environments is commonly approached through a combination of global path planning and local navigation adaptation. Global planning focuses on computing an optimal or near-optimal route between a start and a goal based on a static representation of the environment, typically using graph-based or cost-minimization methods \cite{Hart1968,LaValle2006}. In contrast, local adaptation addresses short-term decision-making under dynamic conditions, such as moving agents, temporary obstacles, or changing environmental constraints, allowing agents to react without recomputing the entire global path \cite{Batty2013}. Hybrid navigation frameworks that integrate global guidance with local, adaptive control are widely adopted in agent-based and geospatial simulation, as they balance optimality, adaptability, and computational efficiency in complex environments \cite{Crooks2008,sutton2018}.

Among global path planning techniques, the A* algorithm remains one of the most widely adopted methods due to its efficiency, optimality guarantees, and suitability for graph-based spatial representations \cite{FOEAD2021507}. For local adaptability, reinforcement learning approaches—most notably Q-learning—have been extensively used to enable agents to adjust their behavior in response to dynamic environmental conditions through trial-and-error interaction \cite{Watkins1992,sutton2018}. The complementary use of A* for global guidance and reinforcement learning for local adaptation has therefore become a common paradigm in agent-based and spatial simulation systems, particularly where environments are partially dynamic but structurally constrained. Recently, deep reinforcement learning approaches, such as Deep Q-Networks (DQN), have been proposed to address navigation problems with high-dimensional state representations by combining Q-learning with deep neural networks \cite{Mnih2015}. While effective in complex perception-driven tasks, such methods typically introduce increased computational cost and reduced interpretability, making simpler tabular reinforcement learning approaches more appropriate for structured, terrain-driven simulation contexts.

\subsection{Terrain Representation for Spatial Analysis and Simulation}
Accurate terrain representation is a foundational requirement for archaeological movement and mobility studies, as elevation, slope, and surface morphology directly affect accessibility, travel cost, visibility, and route selection \cite{Herzog2013LeastCost, Verhagen2019}. Digital Elevation Models (DEMs) have long been used in landscape archaeology to support analyses such as least-cost path modeling, viewshed computation, and catchment analysis, providing a quantitative basis for evaluating how terrain constrains and enables human and animal movement \cite{Conolly2006, Lake2014}. When combined with spatial analysis techniques, DEMs allow the formalization of movement costs associated with slope, distance, and surface difficulty \cite{FLORINSKY201677}. Recent advances in remote sensing have significantly enhanced the quality and resolution of archaeological terrain data. Airborne LiDAR has become a widely adopted method for generating high-precision elevation models, particularly in forested or complex environments where ground visibility is limited \cite{Doneus2011LiDAR, isprs-archives-XLVIII-M-2-2023-357-2023}. Similarly, UAV-based photogrammetry\cite{su13095319} enables the production of detailed surface models through dense image-based reconstruction, offering flexible and cost-effective terrain acquisition at multiple spatial scales \cite{https://doi.org/10.1002/arp.1569}. These techniques capture fine-grained topographic variation and micro-relief features that are often critical for interpreting movement, visibility, and spatial organization in archaeological landscapes. As a result, high-resolution terrain models derived from LiDAR and photogrammetry increasingly underpin contemporary computational approaches to landscape analysis, simulation, and spatial reasoning in archaeology.
Although LiDAR- and photogrammetry-based terrain models provide higher geometric detail, Digital Elevation Models (DEMs) remain widely used in movement and crowd simulation due to their lower computational cost, reduced data complexity, and scalability to large spatial extents. In large-scale simulations spanning tens of square kilometers, very high-resolution terrain data often exceeds the analytical requirements of the research questions while substantially increasing computational demands.
The use of 30 m resolution DEMs from the Copernicus program represents a deliberate balance between spatial realism and analytical tractability. At this resolution, DEMs capture macro-topographic features—such as elevation gradients, slope continuity, ridgelines, and valleys—that exert a primary influence on long-distance human and animal movement. These features are particularly relevant for archaeological analyses concerned with accessibility, travel time, pursuit–evasion dynamics, and transport logistics, where movement is shaped predominantly by sustained slopes and large-scale terrain morphology.
While finer-resolution models better represent micro-topography, they offer limited additional explanatory value for regional-scale mobility analysis and significantly increase computational cost. By employing Copernicus DEMs at 30 m resolution, the framework enables metrically accurate simulation across extensive landscapes while maintaining computational efficiency and model interpretability.

\section{Methodology}
This section describes the methodological framework of the proposed multi-agent simulation, including terrain reconstruction from real elevation data, heterogeneous agent modeling, and a hybrid navigation strategy combining global path planning with local adaptation. These components enable efficient and behavior-aware simulation of movement in uneven archaeological landsc

\subsection{Hybrid Navigation Framework}

Building upon established concepts of global path planning and local navigation adaptation (see Section~2), we implement a hybrid navigation model tailored to dynamic landscapes and heterogeneous agent populations. The proposed approach explicitly incorporates agent-specific mobility characteristics, including walking speed, load, slope tolerance, and behavioral role, ensuring that navigation decisions remain consistent with the modeled agent type.

At the global level, an initial reference path between a start and a goal location is computed using the A* search algorithm on a graph-based representation of the terrain. Traversable spatial units are treated as nodes, while movement costs are weighted according to terrain properties and agent-specific constraints. The A* planner minimizes the evaluation function
\[
f(p) = g(p) + h(p)
\]
where \( g(p) \) denotes the accumulated traversal cost from the start node to node \( p \), and \( h(p) \) is a heuristic estimate of the remaining cost to the goal. This process yields an efficient approximation of a least-cost route under static environmental assumptions. 

However, on high-resolution, large-scale DEMs, the computational overhead of A* is significant. Even on high-performance workstations (e.g., Intel I9 processors, 128GB RAM, and NVIDIA Titan GPUs), initial path calculations for expansive archaeological landscapes can exceed 8–10 minutes. Consequently, recomputing the global A* path every time a dynamic obstacle appears would cause the simulation to stall, making real-time interaction impossible. By introducing local Q-learning, the framework bypasses the need for global replanning, allowing agents to adapt to dynamic changes instantaneously while maintaining the validity of the original long-range path.

To maintain simulation fluidity, local path adaptation is offloaded to tabular Q-learning. Each agent operates in a discrete local state space \( \mathcal{S} \), representing its immediate navigation context, and selects actions \( a \in \mathcal{A} \) corresponding to feasible movement or avoidance maneuvers. The action--value function \( Q(S_t,A_t) \) is updated according to:
\[
Q(S_t,A_t) \leftarrow Q(S_t,A_t) + \alpha \Bigl( R_{t+1} + \gamma \max_{a' \in \mathcal{A}} Q(S_{t+1}, a') - Q(S_t,A_t) \Bigr),
\]
where \( \alpha \) is the learning rate, \( \gamma \) the discount factor, \( R_{t+1} \) the immediate reward, and \( S_{t+1} \) the resulting state. The reward function used for Q-learning is designed to support local path adaptation while preserving consistency with the global A* trajectory. 

Formally, the reward received after executing action \(A_t\) in state \(S_t\) is defined as:
\[
R_{t+1} =
\begin{cases}
- R_{\text{coll}}, & \text{if the agent collides with or cannot bypass an obstacle}, \\[6pt]
- R_{\text{delay}} \cdot \Delta t, & \text{if the agent experiences local navigation delay}, \\[6pt]
- R_{\text{dev}} \cdot d_t, & \text{if the agent deviates from the global A* path}, \\[6pt]
+ R_{\text{rejoin}}, & \text{if the agent successfully rejoins the global path}, \\[6pt]
+ R_{\text{clear}}, & \text{if the obstacle is bypassed efficiently}, \\[6pt]
0, & \text{otherwise}.
\end{cases}
\]

Here, \( d_t \) denotes the distance of the agent from the global A* trajectory at time \( t \), and \( \Delta t \) represents the duration of the current navigation step. The constants \( R_{\text{coll}}, R_{\text{delay}}, R_{\text{dev}}, R_{\text{rejoin}}, \) and \( R_{\text{clear}} \) are scalar weights that control the relative importance of collision avoidance, time efficiency, path adherence, and successful obstacle bypassing.

The interaction between global path planning and local adaptation is formalized through a hierarchical action-selection policy.Let \( P^{*} = \{p_0, p_1, \dots, p_N\} \) denote the global reference path computed by A*, and let \( \chi_t \in \{0,1\} \) be a binary indicator denoting whether the next step along this path is obstructed at time \( t \) by a dynamic obstacle. At time \( t \), the agent tracks the corresponding reference waypoint
\( p_t \in P^{*} \). The agent’s action selection is defined as:
\[
A_t =
\begin{cases}
\arg\min\limits_{a \in \mathcal{A}} \; c_{\text{A*}}(p_t, a), & \text{if } \chi_t = 0, \\[6pt]
\arg\max\limits_{a \in \mathcal{A}} \; Q(S_t, a), & \text{if } \chi_t = 1,
\end{cases}
\]

where \( c_{\text{A*}} \) denotes the local traversal cost toward the next waypoint of the global path and \( Q(S_t,a) \) is the learned action--value function.

\begin{figure}[t]
  \centering
  \includegraphics[width=\linewidth]{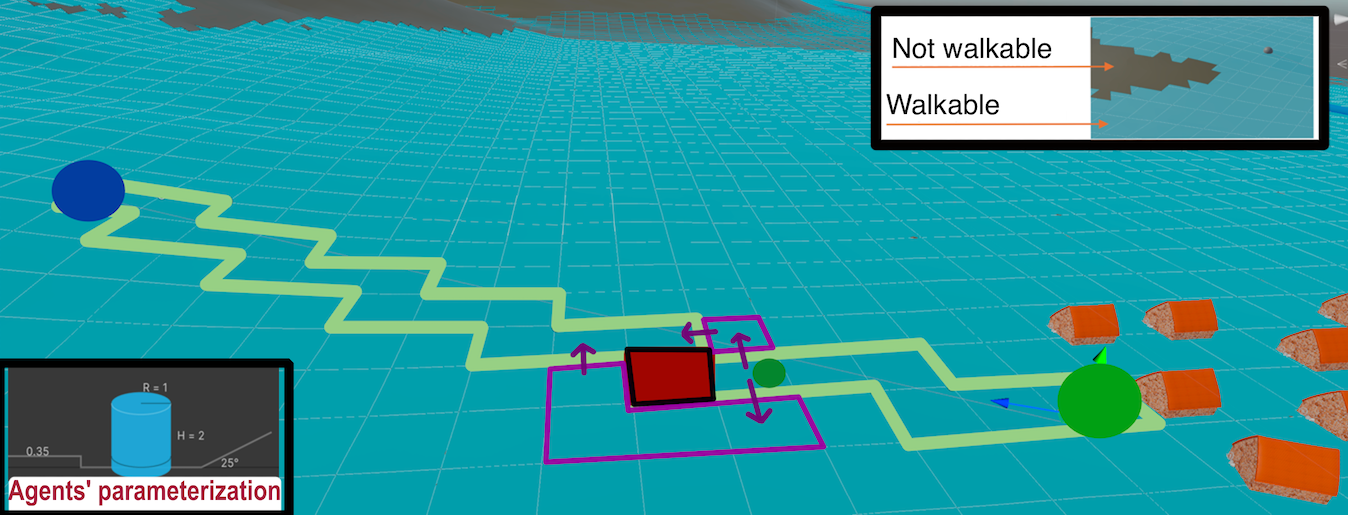}
  \caption{Illustration of the hybrid navigation strategy combining global A* path planning and local Q-learning adaptation.}
  \label{fig:astar-qlearning}
\end{figure}

Figure~\ref{fig:astar-qlearning} illustrates the interaction between global path planning and local adaptation in the proposed navigation framework. The blue circle denotes the target location, while the large green circle indicates the initial position of the agent. The green trajectory represents the global reference path computed using the A* algorithm over the terrain grid. When a dynamic obstacle (red square) obstructs the next segment of this path, local decision-making is activated and the agent (small green circle) temporarily deviates from the global route using Q-learning to select avoidance actions, illustrated by the alternative purple paths. Once the obstacle is bypassed, the agent re-aligns with the original A* path and continues navigation without triggering global replanning.

In the proposed hybrid scheme, A* provides long-range guidance by defining a global reference path, while Q-learning is activated only when dynamic conditions invalidate the next step along this path. The learned local policy is used to bypass the obstruction and guide the agent toward a state compatible with the original A* route. Once this condition is satisfied, control is returned to the global planner without triggering full replanning. Throughout this process, local adaptations remain constrained by the predefined mobility and behavioral features of each agent type.

Deep reinforcement learning was not employed, as the navigation problem involves a limited and structured state space. The combination of A* for global planning and tabular Q-learning for local adaptation provides an efficient, interpretable, and behavior-aware navigation model for the archaeological scenarios examined.

\subsection{Agents modeling}

Agents are modeled with differentiated mobility characteristics to reflect realistic variations in walking behavior across population groups\cite{SUN01031996}. Walking speed is primarily influenced by terrain slope and individual physical condition.

The average walking speed on flat terrain is assumed to be
\[
S_{\text{flat}} \approx 1.42~\text{m/s} \; (5.1~\text{km/h}).
\]

For sloped terrain, a reduction factor \( r \) is applied. For a slope of approximately 15\%, empirical studies suggest a reduction of 30--50\% in walking speed. For example, if we adopt, a  value of 40\%, yielding:
\[
r = 1 - 0.40 = 0.60,
\]
and the resulting slope-adjusted speed is computed as:
\[
S_{\text{new}} = S_{\text{flat}} \times r
\]

\subsubsection{Human-agent modeling}

\begin{figure}[t]
  \centering
  \begin{minipage}[t]{0.389\linewidth}
    \centering
    \includegraphics[width=\linewidth]{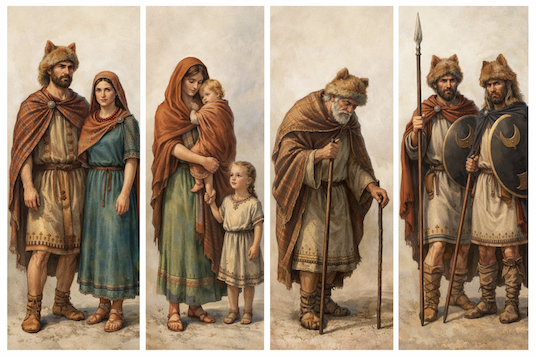}

    \centering\footnotesize \textbf{(a) Human-agent types:} \textit{fit adult, family group, elderly individual and hostile individual.}
  \end{minipage}\hfill
  \begin{minipage}[t]{0.59\linewidth}
    \centering
    \includegraphics[width=\linewidth]{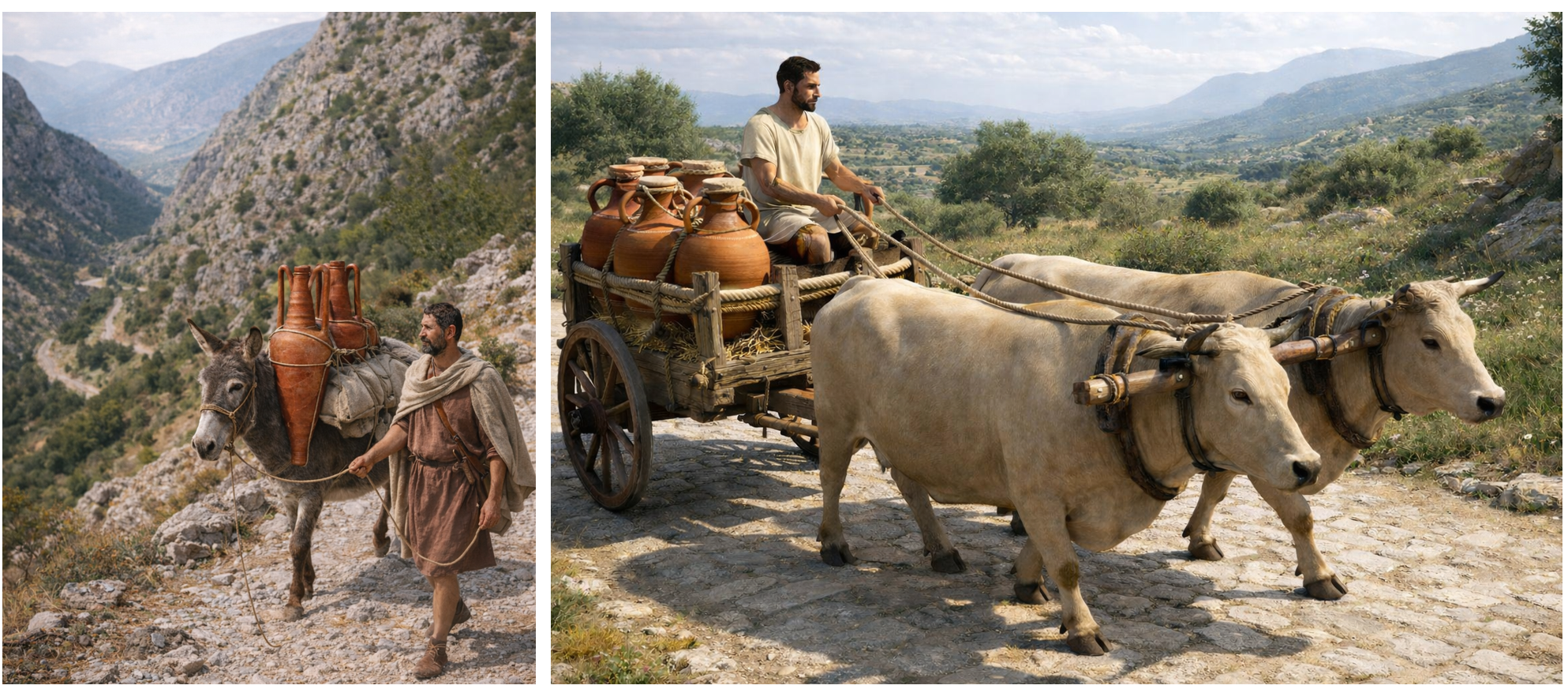}
    
    \centering\footnotesize \textbf{(b) Animal-agent types:} \textit{mule and ox-driven cart.}
  \end{minipage}
  
  \caption{Agent categories modeled in the simulation. Images were generated using AI based on structured scientific text descriptions and visual specifications. Depicted elements such as attire and pottery types were informed by established archaeological evidence and scholarly sources}
  \label{fig:agent-types}

\end{figure}

Human movement in archaeological landscapes is inherently heterogeneous, influenced by physiological, social, and situational factors\cite{SUN01031996}. To capture this variability, human-agents in the proposed multi-agent simulation are modeled as distinct categories, each characterized by different baseline walking speeds and sensitivity to terrain slope. Rather than assuming a uniform population, the model explicitly represents multiple human profiles that are commonly encountered in historical movement scenarios, such as physically \textit{fit adults}, \textit{elderly} individuals, \textit{family} groups with children, and highly motivated or \textit{hostile} individuals, Figure~\ref{fig:agent-types}a. These agent types are subsequently parameterized using empirically grounded speed ranges and reduction factors, allowing the simulation to reflect realistic variations in mobility across uneven terrain. 

At the beginning of the human-agent mobility modeling, a reference terrain slope of 15\% is adopted as it represents a moderate inclination that can be reasonably traversed by most individuals without specialized equipment or training\cite{SUN01031996}. This value is commonly considered an upper threshold for sustained pedestrian movement and allows the simulation to capture realistic mobility constraints while maintaining comparability across agent types. While the terrain slope is kept constant, the reduction factor applied to walking speed is varied across human-agent categories to reflect differences in body type, physical condition, and movement capability. In this way, the model accounts for heterogeneous human mobility by adapting speed reduction parameters to the physiological characteristics and functional abilities of each agent group, rather than assuming a uniform response to terrain difficulty.

Table~\ref{tab:human-agent-mobility} summarizes the mobility parameters assigned to the modeled human-agent groups under a reference terrain slope of 15\%. For each group, the table reports empirically motivated ranges of flat-terrain walking speed and slope-related reduction factors, along with the adopted values used in the simulations and the resulting slope-adjusted speeds. These parameters capture plausible variations in human mobility across different physical conditions and behavioral profiles.

\begin{table}[t]
\centering
\caption{Human-agent mobility parameters under a reference terrain slope of 15\%.}
\label{tab:human-agent-mobility}
\begin{tabularx}{\linewidth}{lXXccX}
\hline
\textbf{Agent type} 
& \textbf{$S_{\text{flat}}$ (m/s)} \newline \small \textit{[Range](Adopted)}
& \textbf{Reduction (\%)}  \newline \small \textit{[Range](Adopted)}
& \textbf{Slope (\%)}
& \textbf{$r$}
& \textbf{$S_{\text{new}}$}  \newline \small \textbf{(m/s) $\thickapprox$ (km/h)} \\
\hline
Fit adults 
& $[1.4, 1.6]\sim(1.5)$ 
& $[20, 30]\sim(25)$
& 15
& 0.75 
& 1.125 $\approx$ 4.0\\

Elderly 
& $[0.8, 1.3]\sim(1.0)$ 
& $[40, 60]\sim(50)$
& 15
& 0.50 
& 0.50 $\thickapprox$ 1.8\\

Families
& $[1.0, 1.3]\sim(1.2)$ 
& $[30, 40]\sim(35)$
& 15
& 0.65  
& 0.78 $\thickapprox$ 2.81\\

Hostile  
& $[1.6, 2.0]\sim(1.8)$ 
& $[10, 30]\sim(20)$
& 15
& 0.80 
& 1.44 $\thickapprox$ 5.18\\
\hline
\end{tabularx}
\end{table}

\subsubsection{Animal-Based Transport Agent Mobility}

In addition to human-agents, we simulated animal-based transport agents that represent historical means of goods transportation, Figure~\ref{fig:agent-types}b. These agents are influenced by both terrain inclination and carried load, which jointly affect their mobility. Walking speed is computed from a baseline flat-terrain speed and adjusted using multiplicative reduction factors.

The general formulation is:
\[
S_{\text{new}} = S_{\text{flat}} \times r_{\text{slope}} \times r_{\text{load}},
\]
where \( r_{\text{slope}} \) and \( r_{\text{load}} \) denote the reduction factors associated with terrain slope and transported load, respectively.


For the examined use cases, two distinct animal-based transport agents are modeled to reflect historically plausible modes of terrestrial movement. Ox-driven cart agents are represented as composite entities consisting of an ox coupled with a wheeled cart carrying four vessels of approximately $100kg$ each, resulting in a total transported load of about $400kg$ (Figure~\ref{fig:agent-types}b, right). In contrast, mule agents are modeled as pack animals transporting two vessels of approximately $50kg$ each, corresponding to a total load of roughly $100kg$ (Figure~\ref{fig:agent-types}b, left). The mobility assumptions, including baseline walking speed, load- and slope-related reduction factors, and resulting effective speed, are summarized in Table~\ref{tab:animal-agent-mobility}. These parameters reflect the fundamental differences between draught-based and pack-based transport, highlighting the greater load capacity but reduced terrain tolerance of ox-driven carts versus the superior adaptability of mules to steeper and more uneven terrain.

\begin{table}[t]
\centering
\small 
\caption{Mobility parameters for animal-based transport agents under slope and load conditions.}
\label{tab:animal-agent-mobility}
\setlength{\tabcolsep}{11pt} 
\begin{tabularx}{\linewidth}{@{} l X X c X c X @{}}
\hline
\textbf{Agent type} 
& \textbf{$S_{\text{flat}}$ (m/s)} \newline\scriptsize\textit{[Range](Adopted)}
& \textbf{Load Red. (\%)} \newline\scriptsize\textit{[Range](Adopted)}
& \textbf{$r_{\text{load}}$}
& \textbf{Slope (\%)}\newline\scriptsize\textit{[Range](Adopted)}
& \textbf{$r_{\text{slope}}$}
& \textbf{$S_{\text{new}}$ \newline\scriptsize(m/s) $\approx$ (km/h)} \\
\hline
Ox-driven cart 
& $[1.0, 1.5] \sim 1.25$
& $[20, 30] \sim 25$
& 0.75
& $[5, 15] \sim 10$
& 0.90
& 0.84 $\approx$ 3.02 \\

Mule 
& $[1.6, 1.8] \sim 1.7$
& $[20, 30] \sim 25$
& 0.75
& $[20, 30] \sim 25$
& 0.75
& 0.96 $\approx$ 3.45 \\
\hline
\end{tabularx}
\end{table}









Although the mule-agents are evaluated under a steeper reference slope of 25\%, their walking speed is not reduced as aggressively, reflecting their ability to traverse narrow, uneven, and rugged terrain with relative agility. In contrast, ox-driven cart agents are assessed under a more moderate slope of 10\%; however, their large body mass, lower maneuverability, and the presence of a wheeled cart impose stronger constraints on movement over irregular surfaces. As a result, the adopted slope-related reduction factors for the two agent types remain relatively close, despite the substantial difference in terrain inclination. This modeling choice captures the distinct locomotion capabilities of pack animals versus draught animals and supports a realistic comparison of alternative transport modes in challenging archaeological landscapes.

\subsection{From Real Elevation Maps to 3D Environment Reconstruction and Simulation}
The simulated environments used in this study are derived from real-world Digital Elevation Models (DEMs) to ensure geometric accuracy and spatial realism. EMs were obtained and processed using \textit{QGIS} and retain their original georeferencing, real-world dimensions, resolution, and elevation values. The elevation data were derived from the Copernicus Digital Elevation Model, based on measurements acquired by the TanDEM-X mission between 2011 and 2015. These DEMs were subsequently transformed into three-dimensional terrain meshes, preserving slope gradients and elevation discontinuities that directly influence agent mobility. The reconstructed terrains were then imported into the game engine, where they form the spatial substrate for agent navigation and interaction.
Preserving real-world scale results in large, high-resolution environments that are computationally demanding. On such terrains, repeated global path recomputation using classical shortest-path algorithms (e.g., A*) alone proved inefficient and unsuitable for real-time simulation. This constraint motivated the adoption of a hybrid navigation strategy that combines global A* planning with local Q-learning-based adaptation, allowing agents to respond to dynamic conditions without costly global replanning (Figure~\ref{fig:astar-qlearning}. The complexity of navigation is further increased by agent-specific physical and mobility parameters—such as body radius, height, load, and maximum traversable slope—which directly affect collision detection, terrain accessibility, and movement costs.
For the Kimmeria case study, the reconstructed terrain covers approximately $16,048 m$ in width and $10,384 m$ in length, encompassing the fortified hilltop and surrounding lowland approaches. Elevation values range from $47.963 m$ to $1,058.686 m$ above sea level, capturing both steep upland relief and gentler slopes. This spatial extent enables the simulation of long-range pursuit–evasion dynamics over realistic distances, where elevation gain, slope persistence, and loss of visibility arise naturally from terrain morphology.
The Kalapodi landscape is similarly reconstructed at full spatial scale, covering approximately $16,515 m$ in width and $10,384$ m in length. Elevation values range from $–2 m$, corresponding to low-lying coastal or near-coastal areas, to $1,058.686 m$, reflecting the mountainous interior through which terrestrial transport routes must pass. Preserving this elevation range is essential for evaluating alternative transport strategies, as slope gradients and route length directly constrain the feasibility of wheeled versus pack-animal movement.
By maintaining real-world spatial dimensions and elevation ranges across both case studies, simulated travel times, path lengths, and movement constraints are expressed in meaningful physical units. This allows simulation outcomes to be interpreted in direct relation to archaeological questions of accessibility, refuge potential, and transport efficiency, while simultaneously justifying the need for an efficient hybrid navigation approach capable of operating on large-scale, high-fidelity terrains.

\subsection{Immersive Agent-Centric Visualization}

\begin{figure}[t]
  \centering
  \includegraphics[width=\linewidth]{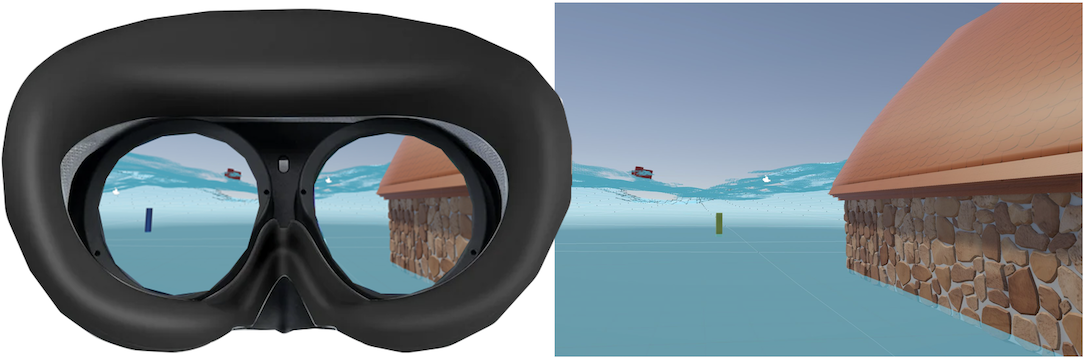}
      \caption{Agent-centric immersive visualization showing the simulated environment from the viewpoint of an agent, enabling inspection of perception, visibility, and interaction dynamics in first-person perspective.}
      
  \label{fig:imersive-view-simulation}
\end{figure}

To support the inspection and interpretation of agent behavior, the proposed framework includes an immersive, Figure \ref{fig:imersive-view-simulation}, agent-centric visualization mode that allows researchers to observe the simulated environment from the perspective of an individual agent. Using a fully immersive head-mounted display, the virtual camera is aligned with the agent’s viewpoint, enabling real-time visualization of what the agent perceives and how it interacts with surrounding terrain, structures, and other entities.

This first-person perspective facilitates qualitative validation of navigation decisions, visibility constraints, and interaction dynamics by allowing domain experts to experience the environment as the agent does. Rather than serving as a user interface or training mechanism, the immersive view functions as an analytical tool that enhances understanding of agent-environment interactions and supports the interpretive assessment of simulation outcomes in complex archaeological landscapes.

\section{Use Cases}

\subsection{Kimmeria: Terrain-Aware Pursuit and Evasion}
\subsubsection{Archaeological question}
\begin{figure}[t]
  \centering
  \includegraphics[width=\linewidth]{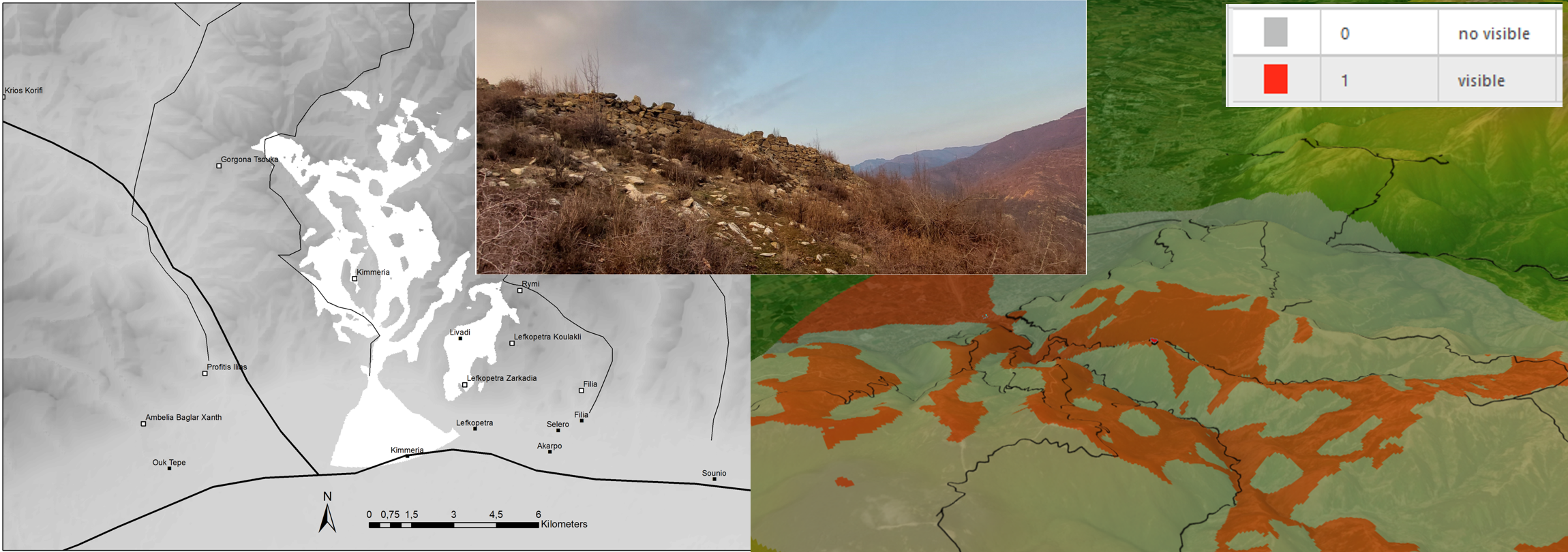}
      \caption{Left-image: The fort at Kimmeria in its broader landscape. Top-image: Remains of the defensive wall, Right-image:Viewshed analysis from the Roman-period fort of Kimmeria, illustrating terrain visibility within an approximate 6 km
radius}
  \label{fig:Kimmeria_1}
\end{figure}

The first case study examines the Roman-period dry-stone fort of Kimmeria, Fig \ref{fig:Kimmeria_1} left image, located at the foothills of the Rhodopi Mountains near modern Xanthi \cite{Triantaphyllos1973Thrace}. Situated at approximately 500 m above sea level, the fort overlooks adjacent lowland settlements as well as mountain corridors leading into the interior of Thrace. The archaeological function of the fort remains debated, with interpretations ranging from military surveillance of the Via Egnatia to route control, ritual use, or use as a local refuge \cite{Evangelidis2021RomeThrace}. The central research question addressed in this case study is whether Kimmeria’s topographic position could have functioned as an effective refuge during episodes of threat, particularly under conditions of active pursuit.

\subsubsection{Limitations of static approaches}
GIS-based analyses, including viewshed (see Fig \ref{fig:Kimmeria_1} -right image) and r.walk modeling, have demonstrated the fort’s strategic visibility and accessibility within the surrounding landscape \cite{Evangelidis2026RomanThrace}. 
However, these approaches are inherently static and cannot account for how movement unfolds dynamically during conflict situations. In particular, they are unable to model simultaneous movement of multiple actors, changes in visibility during pursuit, or the effects of bodily capability and terrain negotiation on escape outcomes.
\subsubsection{Simulation setup}
To address these limitations, a multi-agent-based simulation was implemented using a DEM-derived terrain reconstructed at real-world scale. The scenario models concurrent movement of heterogeneous agent groups across uneven terrain, focusing on pursuit–evasion dynamics between local civilians attempting to reach the fort and hostile agents advancing from the lowlands along established corridors of movement. Agents were assigned terrain-aware mobility parameters reflecting empirically informed differences in walking speed, slope tolerance, and physical constraints.
To capture variability in human mobility, agents were classified into four groups based on age and physical capability: able-bodied adults, elderly individuals, adults accompanied by children, and hostile forces. These categories reflect well-established differences in endurance, speed, and slope negotiation documented in biomechanical and physiological studies. Able-bodied adults represent the upper range of pedestrian mobility, elderly agents were assigned reduced speeds and higher sensitivity to terrain difficulty, and adults with children were modeled with intermediate movement parameters and a preference for less demanding routes. Hostile agents were assigned higher baseline mobility but remained constrained by the same terrain costs.
\subsubsection{Comparison of movement outcomes}
By simulating these agent groups simultaneously under identical environmental conditions, the model evaluates how bodily capability, group composition, and terrain morphology jointly influence accessibility, response times, and pursuit–evasion outcomes. Rather than assuming homogeneous movement behavior, the simulation enables assessment of whether the fort could plausibly have served as an accessible refuge for the broader local population, rather than only for physically capable individuals. This approach highlights how uneven terrain and human heterogeneity can critically shape conflict-related movement dynamics in archaeological landscapes.

\subsection{Kalapodi: Terrestrial Transport Scenarios}
\subsubsection{Archaeological question}
\begin{figure}[t]
  \centering
  \includegraphics[width=1\linewidth]{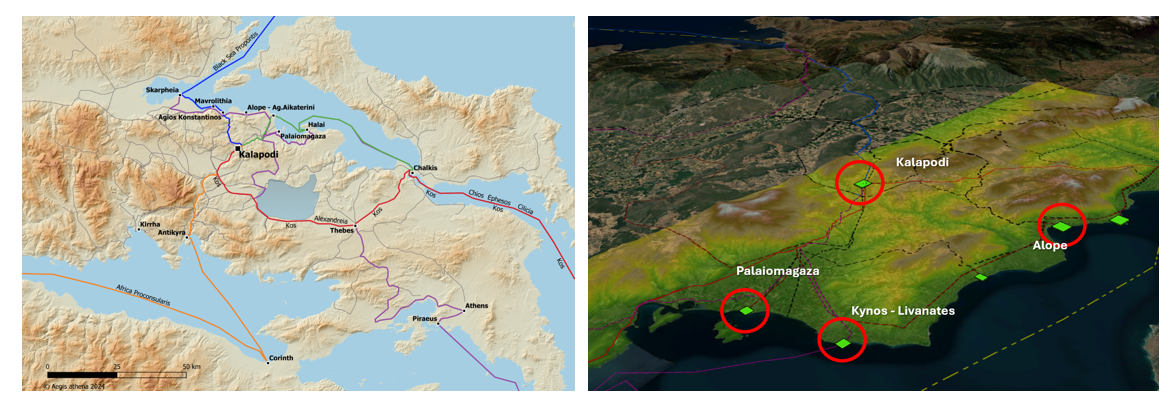}
      \caption{The sanctuary at Kalpodi in its position in local networks (left). Local Harbours in relation to Kalapodi (right)}
  \label{fig:Kalapodi_3}
\end{figure}

The sanctuary of Kalapodi, located in central Greece (see Fig  \ref{fig:Kalapodi_3} left) and systematically excavated by the German Archaeological Institute since 1973 \cite{Niemeier2016Kalapodi, Sporn2016Kalapodi}, functioned as a major religious and socio-economic node from the Late Bronze Age through the Roman period. In the Roman period, the presence of imported ceramics—particularly amphorae—indicates sustained connectivity with wider Mediterranean trade networks \cite{Grigoropoulos2011KeramikDAI,Grigoropoulos2023KalapodiCoarseWares}. The key archaeological question addressed in this case study is how goods were physically transported from coastal entry points to the sanctuary across a rugged inland landscape lacking extensive metalled road infrastructure.

\subsubsection{Limitations of static approaches}
Archaeological, GIS-based, and network analyses suggest that commodities reached Kalapodi via coastal harbors in the North Euboean Gulf, notably Alope, Kynos, and Palaiomagaza (see Fig \ref{fig:Kalapodi_3} right), before being transported inland \cite{GrigoropoulosEvangelidis2026KalapodiNetworks}. While these approaches effectively demonstrate regional connectivity and potential routes, they remain limited in explaining how different transport technologies operated under uneven terrain conditions. In particular, static models cannot capture how slope, load, and transport modality jointly affect travel time, route selection, and carrying efficiency during overland movement.
\subsubsection{Simulation setup}
To address these limitations, a multi-agent-based simulation was implemented within a GIS-derived three-dimensional terrain representing the Kalapodi landscape at real-world scale. Transport agents were defined according to historically documented modes of terrestrial transport: ox-driven carts and pack animals (mules or donkeys). Each agent type was parameterized using values derived from historical sources, experimental archaeology, and biomechanical constraints, including load capacity, baseline speed, and sensitivity to slope and terrain roughness.
\subsubsection{Comparison of transport strategies}.
Ox-driven cart agents were modeled with high carrying capacity but reduced speed and increased sensitivity to slope and surface irregularity, reflecting the constraints of wheeled transport in mountainous environments. Pack-animal agents, by contrast, were assigned lower carrying capacities but greater slope tolerance and more stable movement speeds, enabling navigation across steeper and more irregular terrain. By simulating both transport strategies simultaneously under identical environmental conditions, the model allows direct comparison of travel duration, effective distance, and logistical efficiency. This comparison provides insight into how landscape constraints and transport technology may have favored decentralized, small-load transport systems over bulk movement, shaping the organization and scale of trade associated with the sanctuary of Kalapodi.

\section{Results and Analysis}
\begin{figure}[t]
  \centering
  \includegraphics[width=\linewidth]{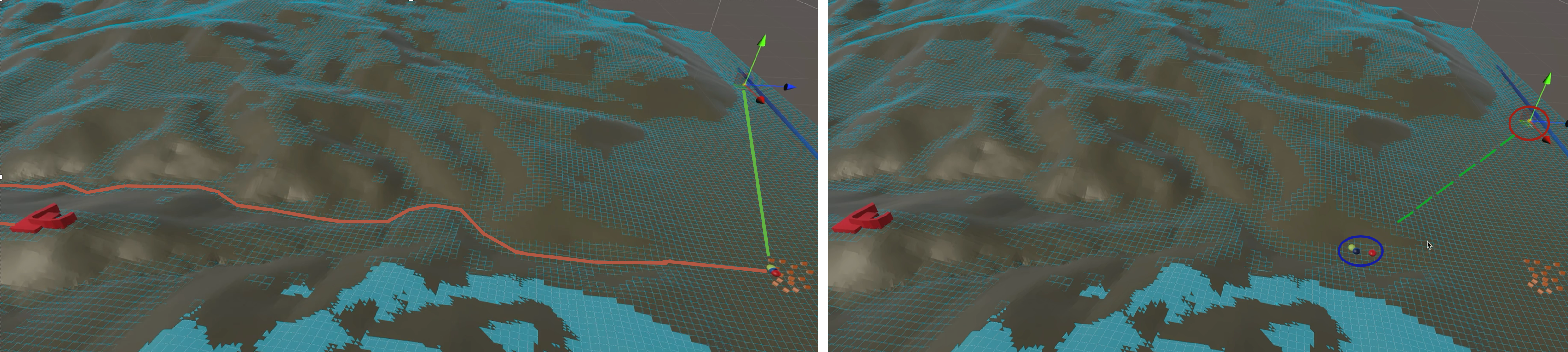}
      \caption{Kimmeria use case showing terrain-aware paths followed by village agents (red) and hostile agents (green). Hillside movement leads to loss of line of sight and termination of pursuit.}
  \label{fig:kimeria-usae-case}
\end{figure}

The simulation results from both case studies demonstrate how terrain morphology, agent heterogeneity, and transport technology interact to shape mobility outcomes in archaeological landscapes—effects that are difficult to capture using static analytical methods alone.
In the Kimmeria case study,  the pursuit–evasion simulations indicate that uneven terrain, combined with time and effort constraints—particularly for the pursuing hostile agents—plays a decisive role in conflict-related movement. As civilian agents ascended toward the fortified hilltop (as illustrated in Fig.~\ref{fig:kimeria-usae-case} leftimage), their trajectories increasingly diverged from direct least-cost routes, instead favoring slopes and terrain features that reduced exposure and disrupted lines of sight. In several simulation runs, loss of contact between pursuing and fleeing agents emerged naturally as a consequence of terrain-induced occlusion, compounded by the increasing energetic cost incurred by the pursuing party (Figure ~\ref{fig:kimeria-usae-case}, right image ) , ultimately leading to the abandonment of pursuit prior to interception.
These outcomes suggest that Kimmeria’s elevated position may have functioned less as a locus of active defense and more as a refuge whose effectiveness was embedded in the surrounding terrain. The ability to exploit slope, concealment, and indirect movement routes appears to have been central to successful evasion, particularly under conditions of active pursuit. Importantly, simulation outcomes varied across agent categories, indicating that bodily capability and group composition significantly influenced accessibility and escape success. This variability supports interpretations of Kimmeria as a refuge whose protective potential would have differed across segments of the local population, rather than as a uniformly accessible defensive stronghold.

The Kalapodi transport simulations reveal markedly different logistical outcomes for alternative terrestrial transport strategies operating under identical environmental conditions. Ox-driven cart agents (Fig. \ref{fig:agent-types}b, right), despite their high carrying capacity—modeled as composite agents transporting four vessels of approximately 100 kg each—experienced substantially longer travel durations. Their reduced speed and heightened sensitivity to slope constrained route selection to gentler terrain, resulting in longer and less direct paths.
In contrast, pack-animal agents (Fig.~\ref{fig:agent-types}b, left) consistently achieved shorter travel times by traversing steeper and more direct routes. These agents were able to operate effectively on slopes of approximately 25

Taken together, these results quantitatively demonstrate how pack-animal transport could outperform wheeled transport in rugged inland landscapes, particularly in regions where metalled road infrastructure was limited or absent. From an archaeological perspective, this supports interpretations that emphasize decentralized, small-load transport systems as a practical and resilient solution for sustaining long-term connectivity between coastal entry points and inland sites such as Kalapodi. Rather than maximizing cargo volume per individual journey, transport efficiency appears to have been achieved through flexibility, greater terrain tolerance, and repeated movement across the landscape.

\begin{figure}[t]
  \centering
  \includegraphics[width=\linewidth]{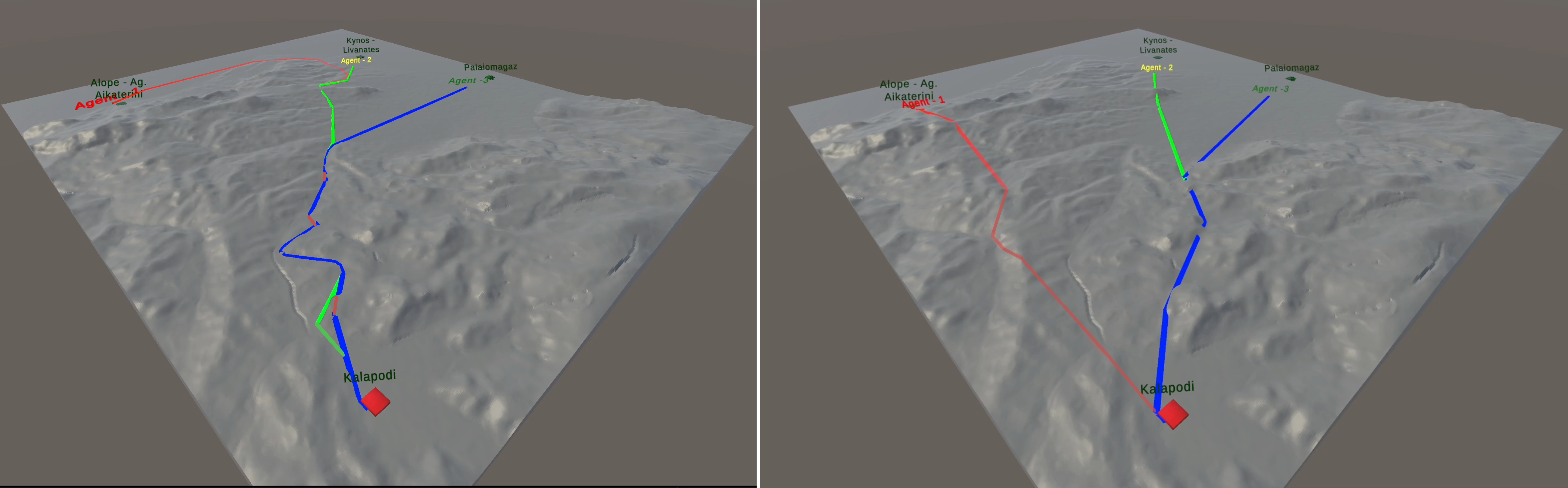}
      \caption{Kalapodi use case showing simulated transport routes for ox-driven cart agents (left) and mule-based agents (right).}
  \label{fig:kalapodi-ox-driven-paths}
\end{figure}
 The mule-based scenario represents pack-animal transport across steeper and more uneven terrain (Figure~\ref{fig:agent-types}b left image). Each mule carries two vessels of approximately 50\,kg each, for a total load of roughly 100\,kg. A higher reference slope of approximately 25\% is considered, reflecting paths unsuitable for wheeled carts but accessible to pack animals. Despite the increased slope, mule agents maintain a higher effective average speed of approximately 0.96\,m/s, resulting in travel durations between roughly 7.5 and 9 hours.

Table~\ref{tab:kalapodi-transport} summarizes the key parameters and outcomes of the two transport scenarios, enabling a direct comparison between alternative movement strategies in the Kalapodi landscape.

\begin{table}[t]
\centering
\caption{Comparison of cart- and mule-based transport simulations for the Kalapodi use case.}
\label{tab:kalapodi-transport}
\begin{tabular}{lccccccc}
\hline
Transport mode & Slope (\%) & Load (kg) & Vessels & Avg. speed (m/s)  \\
\hline
Ox-driven cart & $\sim$10 & $\sim$400 & 4 & $\sim$0.84  \\
Mule & $\sim$25 & $\sim$100 & 2 & $\sim$0.96  \\
\hline
\end{tabular}
\end{table}

Table~\ref{tab:kalapodi-cart-vs-mule} provides a direct comparison between ox-driven cart and mule-based transport for the Kalapodi use case. By evaluating both transport modes across identical origin–destination pairs, the table highlights differences in travel duration and effective distance that emerge from contrasting load capacity and terrain tolerance. The results complement the scenario descriptions by quantitatively illustrating how pack-animal transport can achieve shorter travel times on steeper terrain, despite carrying lighter loads.

\begin{table}[t]
\centering
\small
\caption{Comparison between ox-driven cart and mule transport agents for the Kalapodi use case.}
\label{tab:kalapodi-cart-vs-mule}
\setlength{\tabcolsep}{3pt} 
\begin{tabular}{@{} l l c w{c}{3.2cm} w{c}{3.2cm} w{c}{3.2cm} l @{}}
\hline
\textbf{Agent} & \textbf{Start} & \textbf{End} & \textbf{Cart Duration} & \textbf{Mule Duration} & \textbf{Difference} & \textbf{Reduction. (\%)} \\
\hline
1 & Alope & 
\multirow{3}{*}{\rotatebox{90}{Kalapodi}} 
& $\sim$17:00 h ($\sim$42 km) & $\sim$07:40 h ($\sim$24 km) & $\sim$09:20 h ($\sim$18 km) & $\sim$55.9 (42) \\

2 & Kynos & 
& $\sim$12:30 h ($\sim$31 km) & $\sim$08:50 h ($\sim$28 km) & $\sim$03:40 h ($\sim$2.9 km) & $\sim$29.3 (9) \\

3 & Palaiomagaza & 
& $\sim$12:00 h ($\sim$30 km) & $\sim$08:20 h ($\sim$27 km) & $\sim$03:40 h ($\sim$3.2 km) & $\sim$30.6 (10) \\
\hline
\end{tabular}
\end{table}

Across both case studies, the results highlight the importance of modeling movement as a dynamic, embodied process shaped by landscape constraints and agent-specific capabilities. While the visual representation of terrain in the simulations is intentionally simplified and does not explicitly depict elements such as trees or dense vegetation, these features are incorporated into the navigation and path-planning process as spatial constraints. As a result, agent movement and route selection remain conditioned by environmental affordances and obstacles, even when they are not visually rendered. This design choice prioritizes analytical clarity while preserving the functional influence of landscape features on mobility outcomes.
Taken together, the results demonstrate how agent-based simulation can bridge the gap between static spatial analysis and archaeological interpretation by revealing how movement, visibility, and transport efficiency emerge from the interaction between people, technology, and terrain.

\subsection{Limitations}

Despite its advantages, the proposed framework has several limitations. Although the simulated environments are derived from real-world Digital Elevation Models, the terrain representations are intentionally simplified and do not include explicit modeling of vegetation, built structures, or small-scale surface features. As a result, the simulations focus primarily on topographic constraints such as elevation, slope, and terrain morphology, and do not capture the potential influence of fine-grained environmental or architectural elements on movement and visibility \cite{Landeschi2019RethinkingGIS}. Agent behavior is governed by predefined mobility parameters and localized adaptive learning. Learning is restricted to short-term navigation episodes and does not accumulate across simulations. While this design choice supports interpretability and computational efficiency, it does not model higher-level cognitive, social, or cultural processes, such as long-term learning, memory, cooperation, or collective strategy formation. In addition, the use of tabular Q-learning assumes a limited and structured state space, which restricts scalability to more complex or highly dynamic environments. Finally, mobility parameters are based on empirical approximations and historically plausible estimates rather than direct archaeological measurements, introducing uncertainty in absolute quantitative outcomes. The framework is therefore intended as an exploratory tool, and simulation results should be interpreted as plausible movement scenarios rather than definitive reconstructions of past human behavior.

\section{Conclusions and Future Works}
This paper presented a multi-agent-based modeling framework for investigating archaeological mobility across uneven and data-rich landscapes. By integrating real-world elevation data with agent-specific mobility characteristics, the proposed approach enables the simulation of heterogeneous human and animal movement under terrain, visibility, and interaction constraints. The proposed hybrid navigation framework successfully addresses the computational bottlenecks inherent in high-resolution archaeological simulations. By integrating local RL-based adaptation, we reduced the response time to dynamic obstacles from several minutes—required by global A* replanning. This ensures real-time simulation fluidity without sacrificing the accuracy of long-range mobility patterns in complex, uneven landscapes

Through the Kimmeria and Kalapodi use cases, the framework demonstrated its capacity to explore alternative movement strategies, pursuit–evasion dynamics, and transport efficiency under historically plausible conditions. The results highlight how terrain morphology, load constraints, and agent heterogeneity can significantly influence mobility outcomes that are not easily inferred from static archaeological evidence alone. Additionally, the inclusion of immersive, agent-centric visualization offers a novel analytical perspective, enabling researchers to qualitatively assess perception, visibility, and interaction dynamics directly from the agent’s viewpoint.

Overall, the proposed framework contributes a flexible and interpretable tool for exploratory analysis in archaeological research, supporting scenario-based reasoning rather than deterministic reconstruction and bridging computational modeling with archaeological interpretation.

Future research will address both technical and archaeological challenges to further enhance the realism, expressiveness, and applicability of the proposed framework.

From a technical perspective, a key direction involves modeling more complex and adaptive behaviors. This includes the incorporation of dynamic personality traits that evolve through agent interactions, fatigue and health models that affect performance over time, and emotional or social factors such as trust, cooperation, and conflict. Extending the framework toward multi-agent reinforcement learning will enable the study of coordinated and competitive group behaviors, while pipeline automation will facilitate large-scale experimentation across multiple landscapes and scenarios.

From an archaeological standpoint, future work will expand the framework to maritime contexts, enabling the modeling and analysis of sea-based navigation, trade routes, and naval conflicts using historically grounded environmental and climatic data. Such extensions will allow the investigation of mobility and interaction patterns that transcend terrestrial landscapes and provide a more holistic view of movement, exchange, and conflict in past societies.

Together, these directions aim to advance agent-based simulation as a core methodological tool for archaeological research, supporting richer behavioral modeling, broader spatial domains, and deeper integration between computational methods and archaeological inquiry.

\bibliographystyle{unsrt}  

\bibliography{my-references}

\end{document}